\documentclass[conference]{IEEEtran}
\IEEEoverridecommandlockouts
\usepackage{cite}
\usepackage{amsmath,amssymb,amsfonts}
\usepackage{algorithmic}
\usepackage{acronym}
\usepackage{hyperref}
\usepackage{graphicx}
\usepackage{gensymb}
\usepackage{textcomp}
\usepackage{xcolor}

\renewcommand{\thesection}{\arabic{section}}

\usepackage[tableposition=top]{caption}

\def\BibTeX{{\rm B\kern-.05em{\sc i\kern-.025em b}\kern-.08em
    T\kern-.1667em\lower.7ex\hbox{E}\kern-.125emX}}

\begin{document}

\title{\Large \textbf{Preset-Voice Matching
for Privacy} \\ \textbf{Regulated Speech-to-Speech Translation Systems} }

\author{

\IEEEauthorblockN{\textbf{Daniel Platnick}\IEEEauthorrefmark{1}\textsuperscript{1,}\textsuperscript{2}, \textbf{Bishoy Abdelnour}\IEEEauthorrefmark{1}\textsuperscript{1}, \textbf{Eamon Earl}\IEEEauthorrefmark{1}\textsuperscript{1}, \\ 
\textbf{Rahul Kumar}\textsuperscript{1}, \textbf{Zahra Rezaei}\textsuperscript{1}, \textbf{Thomas Tsangaris}\textsuperscript{1}, \textbf{Faraj Lagum}\textsuperscript{1}}
\IEEEauthorblockA{
\textsuperscript{1}Vosyn, Etobicoke, Canada \\
\textsuperscript{2}Vector Institute, Toronto, Canada\\
\texttt{daniel.platnick@torontomu.ca},\\ \texttt{\{bishoynour,eamon.sc.earl\}@gmail.com}
  }
\thanks{\IEEEauthorrefmark{1}These authors contributed equally and share co-first authorship.}

\thanks{\IEEEauthorrefmark{2}\textit{Accepted to the ACL PrivateNLP 2024 Workshop.}}

}


\maketitle

\begin{abstract}
In recent years, there has been increased demand for speech-to-speech translation (S2ST) systems in industry settings. 
Although successfully commercialized, cloning-based S2ST systems expose their distributors to liabilities when misused by individuals and can infringe on personality rights when exploited by media organizations.
This work proposes a regulated S2ST framework called \emph{Preset-Voice Matching (PVM)}. \emph{PVM} removes cross-lingual voice cloning in S2ST by first matching the input voice to a similar \textit{prior consenting} speaker voice in the target-language. 
With this separation, \emph{PVM} avoids cloning the input speaker, ensuring \emph{PVM} systems comply with regulations and reduce risk of misuse. 
Our results demonstrate \emph{PVM} can significantly improve S2ST system run-time in multi-speaker settings and the naturalness of S2ST synthesized speech. 
To our knowledge, \emph{PVM} is the first explicitly regulated S2ST framework leveraging similarly-matched preset-voices for dynamic S2ST tasks.
\end{abstract}

\section{Introduction}
\label{sec_int}

Progress in deep learning and voice cloning technology has enhanced public access to robust AI-driven voice cloning systems.
These systems can help solve complicated speech-to-speech translation (S2ST) tasks like automated video dubbing (auto-dubbing) by generating audio deepfakes \cite{auto-dubbing, deepfakes-AI, deepfake-good-evil}. 
Cloning systems are desirable for dynamic speech tasks because they can generate a clone from an input voice given an audio sample as short as a few seconds \cite{afew-sample}. 
Currently, voice cloning technology is highly unregulated and can be harmful if misused or commercialized irresponsibly \cite{cloning_problems1_2023}.

As voice cloning systems can clone an arbitrary voice and do not require permission, they raise several privacy concerns \cite{legal-notes}.
Risks related to voice cloning technology include lack of informed consent, biometric privacy, and the spread of misinformation through deepfakes \cite{real-time-deepfake, privacy-concern}. 
Robust regulations are necessary to mitigate these risks, protect individual rights, and prevent misuse \cite{legal-notes, human-right-deepfake, cyber-security}. 

The risks of unregulated voice cloning technologies are compounded by a high demand for voice cloning-based products.
Pressure to capitalize on a newly budding market of cloning-based products can lead businesses to emphasize speed over careful and tested development.
Since voice cloning technology is so new, regulatory measures are required and in the process of being implemented, but not yet fully in place.
Given these challenges, it is crucial to integrate privacy regulations into AI-powered voice cloning systems \cite{protect, prevent-fake,deepfake-metaverse}.

To address the need for regulated voice cloning technology, we propose \emph{Preset-Voice Matching (PVM)}, a regulated S2ST framework. \emph{PVM} bakes regulatory precautions into the S2ST process by removing the explicit objective function of cloning an unknown input speaker’s voice, and instead cloning a \emph{similar} preset-voice of a consenting speaker. \emph{PVM} can be easily installed on top of existing cascaded S2ST pipelines, improving regulatory compliance. We find this process also decreases system run-time in multi-speaker auto-dubbing scenarios and improves speaker naturalness relative to state-of-the-art voice cloning systems when translating across our tested languages.

The intention of this paper is to put forward a regulated \emph{PVM} S2ST framework that is robust against legislative changes and future liability concerns. We demonstrate \emph{PVM} is desirable for S2ST over current benchmark voice cloning frameworks due to its inherent safety, lower run-time in multi-speaker scenarios, and enhanced speaker naturalness. We show this by providing and testing a \emph{PVM} algorithm which we call \emph{GEMO-Match}. 
We hope this work inspires others to develop and tune the framework for different high-performance environments.
Our main contributions are as follows:
\begin{enumerate}
    \item We propose \emph{PVM}, a novel privacy-regulated S2ST framework which leverages consented preset-voices to clone a preset-voice similar to the input voice. 
    \item We provide a gender-emotion based \emph{PVM} algorithm, \emph{GEMO-Match}, and use it to demonstrate \emph{PVM} in multilingual settings.
    \item We empirically analyze \emph{GEMO-Match} in terms of robustness, multilingual capability, and run-time, on two speech emotion datasets and discuss the implications of our system.
    \item We create and provide a \emph{Combined Gender-Dependent Dataset (CGDD)}, which combines various benchmark speech-emotion datasets for training future gender-dependent \emph{PVM} algorithms. 
\end{enumerate}

The rest of this paper is organized as follows. Background information is provided in Section \ref{sec_background}. 
Our \emph{PVM} framework and \emph{GEMO-Match} algorithm are detailed in sections \ref{sec_pvm_framework} and \ref{sec_gemo-match}. 
Relevant datasets are described in Section \ref{sec_dataset_desc}. 
Section \ref{sec_experimental_setup} explains our experimental setup as well as the techniques, algorithms, and parameters used in the study. Section \ref{sec_experimental_results_analysis} includes experimental results and analysis. 
We discuss potential future work towards \emph{PVM} and conclude the paper in sections \ref{sec_future_works} and \ref{sec_conclusion}. 
We address \emph{PVM} limitations in Section \ref{sec_limitations}.

\section{Background Information}
\label{sec_background}

Speech-to-speech translation (S2ST) is typically achieved by direct translation or cascaded approaches \cite{cascade_vs_direct_translation_2022}.
Direct translation approaches use speech and linguistic encoder/decoders \cite{translatotron1_2019} to directly translate speech signals from one language to another.
Cascading architectures split S2ST into three sub-tasks, using separate but connected speech-to-text (STT), text-to-text (TTT), and text-to-speech (TTS) modules \cite{holistic_cascade_2023}. 
Cascading architectures have been the traditional method for S2ST.

Two common approaches for synthesizing speech from text are concatenative and parametric TTS.
Concatenative TTS combines pre-recorded clips from a database to form a final speech output \cite{gujarathi2021review}. 
Parametric TTS attempts to model and predict speech variations given text and a reference voice \cite{parametric_tts_2011}. 
Parametric deep learning methods have shown ubiquitous success spanning various industries from computer vision to text synthesis \cite{deeplearning_ex1, deeplearning_ex2, deeplearning_ex3, deeplearning_ex4}.
As deep neural network (DNN) based TTS methods can lead to natural and expressive synthesized voices, they are desirable for many speech tasks \cite{deep-learning2024}.

Wavenet is a benchmark DNN-based TTS model \cite{oord2016wavenet}. 
Since its creation, there have been many advancements in sequence-to-sequence TTS models trained to produce human-like speech \cite{tacotron1_2022}. 
Wavenet performs speech synthesis by training on a set of human voices, conditioning on their unique speaker ID to generate natural-sounding utterances in the voice of a selected speaker \cite{oord2016wavenet}.
Recently, there have been models which aim to extend this behavior by cloning voices unseen in training, resulting in zero-shot voice cloning \cite{vallex_2023}.

Cross-lingual voice cloning is difficult due to complexities in discriminating between language-specific and speaker-specific features within a singular waveform, and mapping these features across different languages \cite{xtts}.
Additionally, training robust multilingual speech generation models requires vast amounts of processed language and speech data in multiple languages with a variety of utterances and speakers. 
The performance of these models depends on the data they are trained on \cite{data_dependent_models_2017}.

Preset-voice TTS methods generate speech from stored options of preset or pre-recorded voices. 
Preset-voice methods are typically used in static or repetitive systems which do not require dynamic adaptive functionality. 
Examples include pre-programmed transit operator dispatch messages, medical alert systems in healthcare, and emergency flight announcements \cite{bus_dispatch_2001, medical_alert_2013, airport_alert_2019}.

Due to the static nature of current preset-voice methods, they have not previously been used for dynamic S2ST tasks like auto-dubbing.
Such dynamic tasks require modelling different speakers across languages based on incoming media data \cite{auto-dubbing}.
In addition to providing a regulated \emph{PVM} framework, this work aims to extend the application of preset-voice TTS methods to more dynamic settings.

\section{Preset-Voice Matching Framework}
\label{sec_pvm_framework}

This section explains our privacy regulated \emph{Preset-Voice Matching (PVM)} framework.

\emph{PVM} bakes privacy regulations into the S2ST process by cloning a similar and prior consenting preset-voice, instead of the voice originally input to the S2ST system.
The \emph{PVM} framework connects to cascading S2ST architectures, performing additional computations alongside the STT, TTT, and TTS modules. 
The \emph{PVM} framework consists of 3 sub-modules.

Module 1, the \emph{Similarity Feature Extraction} module, extracts features from the inputted voice. 
It then uses the extracted features to match the input voice to the most similar preset-voice from the \emph{Preset-Voice Library}. 
Module 2, the \emph{Preset-Voice Library}, contains a collection of consented target-language preset-voices, partitioned by discrete feature codes depending on the \emph{PVM} implementation. 
Module 3, the \emph{TTS Module}, generates TTS in the target-language using the matched preset-voice from the \emph{Preset-Voice Library}.

We describe these 3 modules below in detail.

\subsection{Feature Extraction and Voice Matching}
The \emph{Similarity Feature Extraction} module extracts meaningful features from the input voice. 
These features are used to determine the most similar consented preset-voice in the target-language from our preset-voice library. 
This module takes in speech signals as input and outputs similarity feature encodings (gender-emotion pair combinations in the case of \emph{GEMO-Match}) to match a consented similar preset-voice. 

\subsection{Target-language Preset-Voice Libraries}
Module 2, the \emph{Preset-Voice Library}, contains a collection of preset-voices in desired target-languages.
The \emph{Preset-Voice Library} acts as a feature codebook, informing the mapping between feature encodings and target-language preset-voice samples.
This module takes in a feature code as input, and outputs a matched consenting speaker preset-voice sample.

\subsection{Text-to-Speech with Matched Preset-Voice}
As input, the \emph{TTS Module} takes in the matched consented preset-voice and target-language text (from an auxiliary TTT module).
The \emph{TTS Module} outputs a clone of the most similar preset-voice in a desired language relative to the features extracted in the \emph{Similarity Feature Extraction} module.
Any voice cloning TTS model supporting the desired target-languages can be used in the \emph{TTS Module}.
Therefore, \emph{PVM} is a general framework and is easily modifiable for many industry settings.

\section{GEMO-Match Algorithm}
\label{sec_gemo-match}
In this section we describe \emph{GEMO-Match}, an example \emph{PVM} framework implementation.

Following a similar notion to \cite{SER_gender_dependent_nn_2023}, \emph{GEMO-Match} employs a hierarchical gender-dependent emotion classifier architecture trained with a gender-dependent training method.
The process of splitting gender and emotion in emotion classification simplifies the emotion classification problem.
As \emph{GEMO-Match} is a \emph{PVM} framework, it contains the 3 \emph{PVM} modules: the \emph{Similarity Feature Extraction} module, the \emph{Preset-Voice Library}, and the \emph{TTS Module}.

These modules and their process are described below.

\subsection{GEMO-Match Modules}

The \emph{GEMO-Match Similarity Feature Extraction} module contains 3 classifiers in two stages. The first stage contains the \emph{gender classifier}, and the second stage includes both the \emph{male-emotion classifier}, and the \emph{female-emotion classifier}. 
The \emph{Similarity Feature Extraction} classifiers are trained in the source language (English).

In \emph{GEMO-Match}, the \emph{Preset-Voice Library} contains previously consenting speakers in desired target-languages for a given S2ST task.
The \emph{Preset-Voice Library} partitions target-language preset-voices by language, gender, and emotion.
The number of target-languages supported by \emph{GEMO-Match} depends on the ability to gather preset-voices in each desired target-language.
The \emph{Preset-Voice Library} in our provided implementation includes two target-languages, French and German.
Therefore, the \emph{GEMO-Match} implementation can translate from English to either French or German.

The \emph{GEMO-Match TTS Module} performs TTS. 
The \emph{TTS Module} is straightforward and performs TTS given a matched preset-voice and a text prompt in the desired target-language. 
We implement \emph{GEMO-Match} with two distinct TTS models, discussed in \ref{subsec_multilingual_setup} and \ref{subsec_runtime_exp}.

\subsection{GEMO-Match Algorithm Flow}

First, source language speech is input to the \emph{Similarity Feature Extraction} module. 
The \emph{gender classifier} then classifies the input voice as male or female. 
Next, given the gender classification result, the source speech is input to the corresponding gender-dependent emotion classifier.
The appropriate gender-dependent emotion classifier will then classify the source language speech as happy, angry, sad, disgust, or neutral.
The two-stage classifier output pair is then concatenated (i.e., Female - Sad). 

The resulting concatenation is used alongside the intended target-language to query the most similar preset-voice in the \emph{Preset-Voice Library}.  
Finally, the feature-matched preset-voice is passed alongside a text prompt to the voice cloning TTS model. 
This algorithm assumes that the intended target-language will be an input to the system.
The performance of \emph{GEMO-Match} depends primarily on the robustness of the \emph{Similarity Feature Extraction} classifiers.

\section{Dataset Descriptions}
\label{sec_dataset_desc}

In this section, we describe the datasets used to test our framework. 

We experimented with two speech-emotion datasets: the Ryerson Audio-Visual Database of Emotional Speech and Song (RAVDESS) \cite{RAVDESS_dataset}, and the \emph{Combined Gender-Dependent Dataset (CGDD)}, which we curated by combining four benchmark speech datasets. 
To ensure compatibility with our gender-emotion based \emph{GEMO-Match} algorithm, we split the RAVDESS dataset by gender and relabeled it with gender-emotion pairs. 
Further details on RAVDESS and \emph{CGDD} are outlined in \ref{subsec_RAVDESS} and \ref{subsec_CGDD}.

\subsection{RAVDESS Dataset}
\label{subsec_RAVDESS}

RAVDESS is a benchmark emotional speech dataset containing 1440 audio files of 24 professional actors (12 female and 12 male) with the emotions calm, happy, sad, angry, fearful, surprise, and disgust \cite{RAVDESS_dataset}. 
As \emph{GEMO-Match} requires consistent labeling across source and target-language data, we focus on a subset of 5 common emotions: happy, angry, sad, disgust, and calm (neutral). 
Each speech sample was originally provided with two intensities, normal and strong.
We filtered the speech files to include only strong intensities as the emotion is more apparent in those samples.
After filtering, the RAVDESS subset contains a total of 5 speech recordings per actor per emotion. 

\subsection{Combined Gender-Dependent Dataset}
\label{subsec_CGDD}

Training a robust gender-emotion classifier requires numerous samples of speakers from various demographics, speaking a variety of utterances with different emotional intensities. 
We found that many available speech-emotion datasets have limited variance in regards to at least one of these features.
To help facilitate gender-dependent training research, we provide a \emph{Combined Gender-Dependent Dataset} \emph{(CGDD)}, made from combining four benchmark emotional speech datasets: RAVDESS, CREMA-D, SAVEE, and TESS \cite{RAVDESS_dataset, CREMAD_dataset, SAVEE_dataset, SP2/E8H2MF_2020}. 

The RAVDESS dataset is explained in section \ref{subsec_RAVDESS}.
CREMA-D is comprised of 7,442 audio recordings of 91 actors. 
These clips include 48 male actors and 43 female actors, with ages ranging from 20 to 74. 
SAVEE database includes four English male speakers aged between 27 and 31, totaling 480 files. 
The TESS database contains two female speakers, one aged 26 and the other aged 64, with a total of 2800 files.  

The \emph{CGDD} dataset is processed for gender-dependent training, useful for hierarchical emotion detection algorithms like \emph{GEMO-Match}.
We further processed the audio based on pitch frequency and loudness to obtain a higher-quality dataset. 
As pitch and loudness are crucial attributes of speech, we filter the data to ensure the files are within a suitable range for speech recognition \cite{pitch_loudness}. 
Additionally, we use RMS loudness to eliminate excessively quiet or loud files.
The best quality was found with a pitch frequency range of 75 Hz to 3000 Hz. 
We removed audio samples with RMS loudness less than -23 dBFS and greater than -20 dBFS.

\subsection{Data Pre-processing}

We processed the RAVDESS and \emph{CGDD} datasets to be compatible with the hierarchical gender-dependent emotion classification architecture of the \emph{GEMO-Match Similarity Feature Extraction} module. 
For both datasets, we partitioned the speech signal files on gender and further organized them into five gender-emotion directories. 
We then converted the speech signals to mel-spectograms using the Fast Fourier Transform. 
Next, the mel-spectrograms were converted to image representation (PNG format) to be processed by a pre-trained ResNet50 model initialized with ImageNet weights \cite{imagenet}.
Our data pre-processing methodology is similar to the procedures outlined in \cite{resnet50_images}.
The Python library Librosa was used to convert speech signal files to mel-spectrogram signals.

\begin{table*}[ht]

\centering
\resizebox{0.7\textwidth}{!}{%
\begin{tabular}{c|cc|cc}
      & \multicolumn{2}{c|}{RAVDESS Precision} & \multicolumn{2}{c}{CGDD Precision} \\
Emotions & Male-Emo          & Female-Emo          & Male-Emo          & Female-Emo          \\ \hline
Happy & 0.78         & 0.56         & 0.51         & 0.78         \\
Angry & 0.78         & 1.00         & 0.82         & 0.87         \\
Sad & 0.50         & 0.40         & 0.59         & 0.66        \\
Disgust & 0.30         & 0.40         & 0.78         & 0.72         \\
Neutral & 0.80         & 0.90         & 0.72         & 0.85         \\  
\end{tabular}
}
\caption{
    Precision of \emph{GEMO-Match} gender-dependent emotion classifiers (ResNet50 pre-trained) on 5 emotions from RAVDESS and \emph{CGDD}. Training the ResNet50 on the \emph{CGDD} dataset results in better generalization across emotions in terms of precision.
}
\label{tbl:emo_classifier_precision}
\end{table*}

\section{Experimental Setup}
\label{sec_experimental_setup}

This section details the setup of each experiment, which show additional strengths of the \emph{PVM} framework, beyond its inherent regulatory benefits. 

We demonstrate the effectiveness of \emph{PVM} for S2ST with \emph{GEMO-Match} in terms of robustness, multilingual capability, and run-time. 
Our experiments were run on a single Tesla T4 GPU with 40 cores.
We discuss each experiment in detail below.

\subsection{GEMO-Match Robustness}
\label{subsec_GEMO-Match_robustness_setup}

For this test, we assess the robustness of \emph{GEMO-Match}. 
The performance of \emph{GEMO-Match} depends on the three \emph{Similarity Feature Extraction} classifiers. 
We fine-tuned and evaluated these classifiers on the RAVDESS and our \emph{CGDD} dataset in terms of accuracy and precision. 
Each classifier was implemented as a ResNet50 previously pre-trained on ImageNet. 
The results of the six classifiers are shown in tables \ref{tbl:emo_classifier_precision} and \ref{tab:emotion_classifier_acc}.

The same approach was used to train each ResNet50. 
The \emph{gender classifiers} were trained for 20 epochs, while the \emph{male-emotion} and \emph{female-emotion classifiers} required 30 epochs to converge. 
Each emotion classifier was trained using a dynamic learning rate schedule: 0.01 for the first 20 epochs, reduced to 0.001 for the remaining 10. 

We used the Adam optimizer, and the Pytorch ImageDataGenerator function for data augmentation \cite{adam_op_2017}. 
The classifiers were trained using a batch size of 32 and a train-test-validation split of 60-20-20. 
The models were optimized using categorical cross entropy as the loss function, incorporating batch normalization and dropout layers for regularization. 
The activation functions used were ReLU for internal layers and softmax for the output layer.

\subsection{GEMO-Match Multilingualism}
\label{subsec_multilingual_setup}

We test \emph{GEMO-Match} in terms of speaker naturalness on the task of translating English speech into French and German speech.
\emph{GEMO-Match} is implemented within a cascaded S2ST system using SeamlessM4T for TTT, and XTTS as the TTS module \cite{communication2023seamlessm4t,xtts}. XTTS is a state-of-the-art TTS model which supports zero-shot voice cloning across 17 languages.
Instead of performing STT, we provide ground truth source-language (English) text directly to the TTT model (SeamlessM4T) to measure the isolated performance of \emph{GEMO-Match} across multiple languages.
We measured speaker naturalness using the standard metric Non-intrusive Objective Speech Quality Assessment (NISQA) \cite{NISQA_2021, NISQA2_2022}. 

We show \emph{PVM} algorithms lead to higher naturalness in S2ST outputted speech by alleviating the need to perform \textit{cross-lingual} voice cloning. 
We compare two cases of S2ST. 
The first case is when XTTS performs cross-lingual cloning from an English voice input to the target-languages German and French. 
In the second case, \emph{GEMO-Match} performs the cross-lingual matching, allowing XTTS to run monolingual TTS given the matched target-language voice as input. 

The French and German preset-voices used in this experiment are sourced from the CAFE, and EmoDB datasets respectively \cite{cafe, emodb}. 
For each target-language in both experimental pipelines, we used 150 English text transcriptions from the CREMA-D dataset alongside emotive English audios from RAVDESS as input \cite{CREMAD_dataset}. We ensured that our RAVDESS audios had an average NISQA (3.54) similar to the preset-voices in our target-languages. For additional context, we included the average preset-voice NISQA scores for both target-languages in Table \ref{tab:multilingual}.  

\subsection{GEMO-Match Run-time}
\label{subsec_runtime_exp}
We compared the run-time of \emph{GEMO-Match} to state-of-the-art TTS models VALL-E X, XTTS, SeamlessM4T, and OpenVoice, as shown in Figure \ref{experiment_runtime} \cite{openvoice_2024}. 
The \emph{gender}, \emph{male-emotion}, and \emph{female-emotion classifiers} were implemented using the same lightweight ResNet50 models as in \ref{subsec_GEMO-Match_robustness_setup}.
Each model was given 10 identical utterances with their respective transcriptions, and average inference run-times were calculated. 
The inputs were each 15 seconds and varied in tone, emotion, pacing, and vocabulary. 

We compared \emph{PVM} (using \emph{GEMO-Match}) with OpenVoice as they are both cascaded TTS frameworks that decouple voice-cloning from voice synthesis. 
OpenVoice uses a variation of VITS for TTS in its open-source implementation \cite{vits}. 
For consistent comparisons with OpenVoice, we use StyleTTS2 for TTS with \emph{GEMO-Match} \cite{styletts}.
StyleTTS2 and VITS are both styling-based models and display similar run-times.
StyleTTS2 is a monolingual TTS model, and we use it to show the run-time benefits of \emph{PVM} removing cross-lingual voice cloning in cascaded S2ST systems.

Figure \ref{fig_sequential_inference} compares \emph{GEMO-Match} with the OpenVoice framework in terms of run-time scaling in multi-speaker scenarios.
We plotted the number of times each system must re-run auxiliary modules while performing TTS over time in multi-speaker instances.
The plots were generated using Python.

\section{Experimental Results and Analysis}
\label{sec_experimental_results_analysis}

In this section, we discuss and analyze our experimental results. 

Section \ref{subsec_robustness} describes the results of the \emph{GEMO-Match} robustness experiment, contained in tables  \ref{tbl:emo_classifier_precision} and \ref{tab:emotion_classifier_acc}. Next, section \ref{subsec_multilingual_results} provides an analysis on the results in Table \ref{tab:multilingual}. Section \ref{subsec_runtime_analysis} then highlights our run-time experiment results.

\begin{table}
  \centering
  \resizebox{0.45\textwidth}{!}{%
  \begin{tabular}{c|c|c}
     & \multicolumn{1}{c|}{RAVDESS} & \multicolumn{1}{c}{CGDD} \\
    Classifier & Accuracy & Accuracy \\
    \hline
    Gender & 94.00  & \textbf{97.00} \\
    Male-Emotion & 62.00 & \textbf{63.21} \\
    Female-Emotion  & 65.00 & \textbf{71.29} 
  \end{tabular}
  }
  
  \caption{Test set accuracies of \emph{GEMO-Match} classifiers.} 
  \label{tab:emotion_classifier_acc}
\end{table}

\subsection{GEMO-Match Robustness Results}
\label{subsec_robustness}

Tables \ref{tbl:emo_classifier_precision} and \ref{tab:emotion_classifier_acc} show the precision and accuracy of the \emph{Similarity Feature Extraction} module classifiers.
Testing \emph{GEMO-Match} on RAVDESS across emotions, the \emph{Male-Emotion Classifier} performs best on happy, angry, and neutral, which have precision scores of 78\%, 78\%, and 80\%, respectively.
The \emph{Female-Emotion Classifier} performs well on angry and neutral, achieving 100\% and 90\% precision, respectively.
We find \emph{GEMO-Match} overfits to certain gender-emotion classes when trained on RAVDESS. 
This is prevalent in the \emph{Female-Emotion Classifier} performance, as it classifies angry emotions with perfect precision, but classifies sad and disgust with 40\% precision.

As illustrated in Table \ref{tbl:emo_classifier_precision}, \emph{GEMO-Match} generalizes more consistently across emotions when trained on \emph{CGDD} compared to \emph{RAVDESS}.
In the cases of both datasets shown in Table \ref{tbl:emo_classifier_precision}, \emph{GEMO-Match} tends to classify angry and neutral effectively.
The improvements in generalization described in Table \ref{tbl:emo_classifier_precision} when using \emph{CGDD} instead of RAVDESS showcases that some benchmarks are currently lacking variation.
\emph{CGDD} can remedy this, as it has higher variance compared to RAVDESS, comprising of multiple benchmark datasets as described in section \ref{subsec_CGDD}.

Table \ref{tab:emotion_classifier_acc} shows the accuracy of \emph{GEMO-Match} on RAVDESS and \emph{CGDD}. 
The \emph{GEMO-Match} gender classifier scored 94\% accuracy on the RAVDESS dataset, and 97\% on \emph{CGDD}.
The best \emph{GEMO-Match} emotion classifier results are found when training and testing on \emph{CGDD}, which results in 63\% accuracy for the \emph{Male-Emotion Classifier} and 71\% for the \emph{Female-Emotion Classifier.} 
Therefore, our proposed \emph{CGDD} dataset can improve model generalization compared to benchmark datasets like RAVDESS.

\subsection{GEMO-Match Multilingual Results}
\label{subsec_multilingual_results}

The results in Table \ref{tab:multilingual} show \emph{PVM} implementations can significantly improve the output naturalness of S2ST systems by enabling monolingual TTS within S2ST. 
We find this trend holds across the two tested languages, French and German.
When XTTS performs cross-lingual TTS from English to German, NISQA values decrease from 3.54 (English) to 3.41 (German).
Similarly, when XTTS cross-lingually clones from English to French, the  input-output NISQA values are 3.54 and 3.54, respectively.
Overall, XTTS either maintained or degraded the input naturalness when performing cross-lingual cloning in our experiments.

We find XTTS performs much better in a monolingual setting, which can significantly enhance S2ST quality.
The average NISQA score when XTTS cloned from German preset-voices to German outputs increased from 3.47 to 3.69. 
The same increase is seen with French, though to a lesser degree.
For our tested language pairs, \emph{GEMO-Match} consistently improves output naturalness by allowing S2ST pipelines to clone in a monolingual context while maintaining cross-lingual behavior.

\subsection{GEMO-Match Run-time Results}
\label{subsec_runtime_analysis}

\begin{figure}
  \includegraphics[width=\columnwidth]{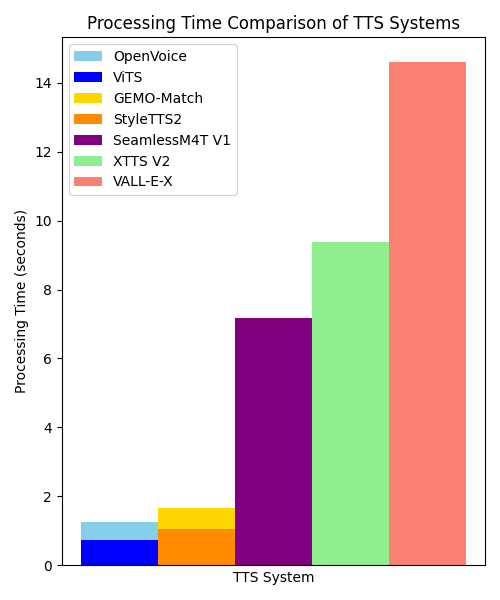}
  \caption{Comparative processing times of different models. OpenVoice's \emph{tone extractor} and \emph{GEMO-Match} are distinguished from their TTS processing times.}
  \label{experiment_runtime}
\end{figure}

The run-time results of different TTS approaches are shown in Figure \ref{experiment_runtime}.
VALL-E X and XTTS, deep multilingual voice cloning models, are slowest on average.
SeamlessM4T offers multilingualism in multiple modalities, but does not clone voices, and has significantly lower runtime than the aforementioned models.
This underscores additional complexities inherent to achieving speech translation and voice cloning in a single embedding space.

In our experiments, the lowest run-times were achieved by our \emph{PVM} implementation (\emph{GEMO-Match} with StyleTTS2) and OpenVoice. Both of these frameworks are not strictly limited to a specific TTS module for processing.
As such, the runtime of their auxiliary, decoupled systems are noted separately in Figure \ref{experiment_runtime}. 
OpenVoice uses the post-processing tone extractor described in \cite{openvoice_2024}, and \emph{PVM} uses \emph{GEMO-Match}. 
For these isolated auxiliary modules, we achieved an average runtime of 0.52 for OpenVoice and 0.61 seconds for \emph{GEMO-Match}.

\begin{table}
  \centering
  \resizebox{0.45\textwidth}{!}{%
  \begin{tabular}{c|c|c}
    \textbf{Target} & \textbf{XTTS Input} & \textbf{XTTS Output} \\ 
    \textbf{Language} & \textbf{NISQA} & \textbf{NISQA} \\ \hline
    \multicolumn{3}{l}{\textit{Cross-lingual Cloning (English prompt)}} \\ \hline
    French & 3.54 & 3.54 \\
    German & 3.54 & 3.41 \\ \hline
    \multicolumn{3}{l}{\textit{Monolingual Cloning (PVM-matched preset)}} \\ \hline
    French & \textbf{3.39} & \textbf{3.43} \\
    German & \textbf{3.47} & \textbf{3.69} \\ \hline
  \end{tabular}
  }
  \caption{Speech quality behavior when decoupling multilingual transformation and voice cloning in S2ST. XTTS performs significantly better when cloning in a monolingual context. Inputs are shown in parentheses.}
  \label{tab:multilingual}
\end{table}

\begin{figure}
  \includegraphics[width=\columnwidth]{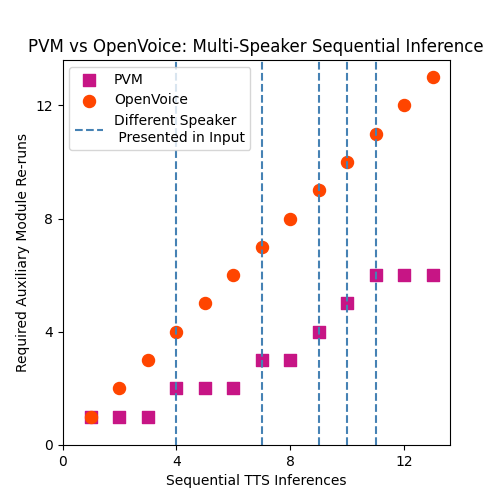}
  \caption{The OpenVoice tone extractor post-processes every TTS output. \emph{GEMO-Match} only needs to re-run on the arrival of a different speaker from the one present in the previous input. }
  \label{fig_sequential_inference}
\end{figure}

Figure \ref{fig_sequential_inference} compares these auxiliary modules under sequential inference on long multi-speaker inputs. 
For this comparison, we focus on the run-time of the entire S2ST system.
Figure \ref{fig_sequential_inference} shows that \emph{GEMO-Match} need only run when a new speaker is presented in the input, while OpenVoice must always post-process the TTS output to achieve the desired result. 
Therefore, \emph{PVM} offers favourable scaling properties, making it desirable for many commercial use-cases.

\section{Future Work}
\label{sec_future_works}

\emph{PVM} is a general framework for regulated S2ST that can be integrated into pre-existing cascaded S2ST pipelines.
The performance of \emph{PVM} is directly dependent on the quality of the individual swappable components of the pipeline.
Consequently, the efficacy of any \emph{PVM} implementation is expected to increase with general advancements in TTS technology.
There are many ways to improve the \emph{PVM} framework, and we propose some ideas for future work.

For future work, we propose a cascaded voice cloning TTS system which uses an initial \textit{vocal encoder} with learned weights to extract and compress relevant features from the input voice. 
The system would perform the classical cloning tasks on this encoded voice in a downstream, decoupled TTS model. 
This would allow voices to be stored in the \emph{Preset-Voice Library} in their encoded formats rather than speech signals, likely decreasing run-time complexity.
Using a cascaded learning process, the TTS module would learn to effectively clone and only synthesize voices encoded by the \textit{vocal encoder}. 
During distribution of the system, the \emph{vocal encoder} would not be published.
In this way, the system could not be used to clone a voice outside of the pre-encoded preset-voices in the \emph{Preset-Voice Library}.

\emph{GEMO-Match} uses classifiers which depend on labeled data.  
This dependency motivates the development of alternative \emph{PVM} instances capable of voice-matching without relying on labeled data. 
We posit that learned encodings can be used, akin to self-supervised learning mechanisms employed by transformer architectures, to extract robust internal representations of speech inputs \cite{devlin2019bert, babu2021xlsr}. 
This would require a new training pipeline with an objective function for maximizing speaker similarity between the input voice and the matched voice. 
The resulting \emph{PVM} system could use latent feature representations to perform voice matching, and training would not require labeled speech datasets.

\section{Conclusion}
\label{sec_conclusion}

We proposed \emph{Preset-Voice Matching}, a novel framework that bakes regulatory precautions into the S2ST process. 
\emph{PVM} achieves this by removing the explicit objective of cloning an unknown input speaker’s voice, and instead cloning a similar preset-voice of a consenting speaker. This paradigm is extensible to a variety of industry settings to regulate the behavior of S2ST systems.
Quantitative experiments show \emph{PVM} is a desirable framework compared to the tested benchmarks in terms of run-time and naturalness of multilingual translation output. 
Additionally, we provided \emph{CGDD}, a gender-dependent speech-emotion dataset. 
We then showed \emph{CGDD} leads to better model generalization and robustness in terms of accuracy and precision compared to the benchmark RAVDESS dataset.
We hope this work inspires others to create more privacy regulated S2ST systems using the \emph{PVM} framework.

\section{PVM Limitations}
\label{sec_limitations}

In this section, we discuss the limitations of \emph{GEMO-Match} and the \emph{PVM} framework.

\emph{GEMO-Match} requires training 3 unique classifiers for every source-language supported by the system.
Specifically, the three \emph{Feature Extraction Module} classifiers need to be trained on language specific emotional speech datasets processed into 3 versions: the entire dataset labeled by gender, and two subsets containing the gender-dependent labeled data.
Gathering and processing data as described for each desired source-language may be complicated depending on data availability.

We acknowledge that the three features language, gender, and emotion alone are inadequate to fully capture the breadth of speaker variance across human speech. 
There are scenarios which demand more closely matched consented speakers in terms of vocal characteristics of the input speaker.
\emph{GEMO-Match} has strong limitations in this respect, which necessitates systems with more granularity in terms of speech feature extraction than what is offered by \emph{GEMO-Match}.

Additionally, \emph{PVM} makes no attempt to mimic background ambience or environmental noise in the inputted audios, as it loses this information when matching to a preset-voice. 
This is a drawback of \emph{PVM}, as maintaining background audio noise information is highly important in some settings. 
However, many modern S2ST systems denoise input audio to improve model performance, and add the noise back during post-processing. \emph{PVM} would not be limited in such an environment, and can ensure high-quality voice inputs to the TTS module by always mapping to high-quality consenting speaker audios.

Lastly, we consider the drawback of error propagation in the \emph{PVM} framework, inherent to cascaded architectures with separate modules. Ultimately, using a set of separate modules introduces additional points of failure, causing inaccuracies which are passed to downstream tasks.

\section{Appendix}

\subsection{Industry Applications}

In this section, we include some examples of cases where \emph{PVM} can be applied to industry settings.

APIs are a common avenue for controlled public access to ML models and pipelines. These access points are commonly subjected to adversarial attacks, where imperceptible artefacts are injected into inputs to produce undesirable results. In the \emph{PVM} framework, the audio input by our user is not directly passed to the TTS model, and is only matched to a consented speaker using feature similarity. This limits the scope of poor results that could be triggered by an adversarial user by negating direct access to the TTS model.
Additionally, propagating audio input data from a genuine user through fewer modules in the pipeline limits opportunities for sensitive bio-metric data to be extracted by malicious third parties.
Ultimately, removing direct control over synthesis of the input voice prevents bad actors from cloning a non-consenting speaker for nefarious goals.

We also consider how \emph{PVM} can be extended to help regulate open-source models. 
As mentioned in Section \ref{sec_future_works}, an autoencoder could be applied to derive robust latent space representations of the preset-voices. 
Matching based on similarity would still occur on the raw preset-voice audios, but their corresponding preset encodings would be passed as input to the voice cloning TTS model. 
The encoder/decoder models would not be published alongside the rest of the system. 
As the TTS model would have only been trained on the latent embeddings, the published system could not be hijacked to clone non-consenting voices.

In content localization systems, media content is leased by distributing platforms, while rights to the reproduction of the likenesses of individuals present in the content is not readily available. 
Not only can \emph{PVM} secure these systems in the manners mentioned above, but its regulated application can help bring this budding market to life by efficiently producing translated content in only the voices of consenting speakers. 
We believe \emph{PVM} provides feasibility to the commercialization of such systems while being robust against future industry regulations.

We hope these examples give insight into the vast extensibility of the \emph{PVM} framework.

\section{Acknowledgements}
We thank the anonymous reviewers for their feedback on this work. 
We also thank Joy Christian, Chao-Lin Chen, Sina Pordanesh, Akash Lakhani, Kaivil Brahmbhatt, and Darshan Sarkale of Vosyn's PVM team for their contributions on the early stages of this work.

\bibliographystyle{IEEEtran}
\bibliography{references}

\begin{thebibliography}{10}
\providecommand{\url}[1]{#1}
\csname url@samestyle\endcsname
\providecommand{\newblock}{\relax}
\providecommand{\bibinfo}[2]{#2}
\providecommand{\BIBentrySTDinterwordspacing}{\spaceskip=0pt\relax}
\providecommand{\BIBentryALTinterwordstretchfactor}{4}
\providecommand{\BIBentryALTinterwordspacing}{\spaceskip=\fontdimen2\font plus
\BIBentryALTinterwordstretchfactor\fontdimen3\font minus \fontdimen4\font\relax}
\providecommand{\BIBforeignlanguage}[2]{{%
\expandafter\ifx\csname l@#1\endcsname\relax
\typeout{** WARNING: IEEEtran.bst: No hyphenation pattern has been}%
\typeout{** loaded for the language `#1'. Using the pattern for}%
\typeout{** the default language instead.}%
\else
\language=\csname l@#1\endcsname
\fi
#2}}
\providecommand{\BIBdecl}{\relax}
\BIBdecl

\bibitem{auto-dubbing}
W.~Brannon, Y.~Virkar, and B.~Thompson, ``Dubbing in practice: A large scale study of human localization with insights for automatic dubbing,'' 2022.

\bibitem{deepfakes-AI}
M.~R. Shoaib, Z.~Wang, M.~T. Ahvanooey, and J.~Zhao, ``Deepfakes, misinformation, and disinformation in the era of frontier ai, generative ai, and large ai models,'' 2023.

\bibitem{deepfake-good-evil}
\BIBentryALTinterwordspacing
N.~Amezaga and J.~Hajek, ``Availability of voice deepfake technology and its impact for good and evil,'' in \emph{Proceedings of the 23rd Annual Conference on Information Technology Education}, ser. SIGITE '22.\hskip 1em plus 0.5em minus 0.4em\relax Association for Computing Machinery, 2022. [Online]. Available: \url{https://doi.org/10.1145/3537674.3554742}
\BIBentrySTDinterwordspacing

\bibitem{afew-sample}
S.~O. Arik, J.~Chen, K.~Peng, W.~Ping, and Y.~Zhou, ``Neural voice cloning with a few samples,'' 2018.

\bibitem{cloning_problems1_2023}
C.~Liu, J.~Zhang, T.~Zhang, X.~Yang, W.~Zhang, and N.~Yu, ``Detecting voice cloning attacks via timbre watermarking,'' 2023.

\bibitem{legal-notes}
A.~Baris, ``Ai covers: legal notes on audio mining and voice cloning,'' in \emph{AI covers: legal notes on audio mining and voice cloning}, 2024.

\bibitem{real-time-deepfake}
G.~Frankovits and Y.~Mirsky, ``Discussion paper: The threat of real time deepfakes,'' 2023.

\bibitem{privacy-concern}
C.~Okolie, ``Artificial intelligence-altered videos (deepfakes) and data privacy concerns,'' \emph{Journal of International Women's Studies}, vol.~25, p.~13, 03 2023.

\bibitem{human-right-deepfake}
F.~R. Moreno, ``Generative ai and deepfakes: a human rights approach to tackling harmful content,'' \emph{International Review of Law, Computers \& Technology}, vol.~0, no.~0, pp. 1--30, 2024.

\bibitem{cyber-security}
K.~N. Sudhakar and M.~Shanthi, ``Deepfake: An endanger to cyber security,'' in \emph{2023 International Conference on Sustainable Computing and Smart Systems (ICSCSS)}, 2023, pp. 1542--1548.

\bibitem{protect}
\BIBentryALTinterwordspacing
Z.~Liu, Y.~Zhang, and C.~Miao, ``Protecting your voice from speech synthesis attacks,'' in \emph{Proceedings of the 39th Annual Computer Security Applications Conference}, ser. ACSAC '23.\hskip 1em plus 0.5em minus 0.4em\relax Association for Computing Machinery, 2023, p. 394–408. [Online]. Available: \url{https://doi.org/10.1145/3627106.3627183}
\BIBentrySTDinterwordspacing

\bibitem{prevent-fake}
W.~J. Tee and R.~Murugesan, \emph{Protecting Data Privacy and Prevent Fake News and Deepfakes in Social Media via Blockchain Technology}.\hskip 1em plus 0.5em minus 0.4em\relax Springer, 02 2021, pp. 674--684.

\bibitem{deepfake-metaverse}
\BIBentryALTinterwordspacing
S.~Tariq, A.~Abuadbba, and K.~Moore, ``Deepfake in the metaverse: Security implications for virtual gaming, meetings, and offices,'' in \emph{The 2nd Workshop on the security implications of Deepfakes and Cheapfakes}.\hskip 1em plus 0.5em minus 0.4em\relax ACM, Jul. 2023. [Online]. Available: \url{http://dx.doi.org/10.1145/3595353.3595880}
\BIBentrySTDinterwordspacing

\bibitem{cascade_vs_direct_translation_2022}
\BIBentryALTinterwordspacing
T.~Etchegoyhen, H.~Arzelus, H.~Gete, A.~Alvarez, I.~G. Torre, J.~M. Martín-Doñas, A.~González-Docasal, and E.~B. Fernandez, ``Cascade or direct speech translation? a case study,'' \emph{Applied Sciences}, vol.~12, no.~3, 2022. [Online]. Available: \url{https://www.mdpi.com/2076-3417/12/3/1097}
\BIBentrySTDinterwordspacing

\bibitem{translatotron1_2019}
Y.~Jia, R.~J. Weiss, F.~Biadsy, W.~Macherey, M.~Johnson, Z.~Chen, and Y.~Wu, ``Direct speech-to-speech translation with a sequence-to-sequence model,'' 2019.

\bibitem{holistic_cascade_2023}
W.-C. Huang, B.~Peloquin, J.~Kao, C.~Wang, H.~Gong, E.~Salesky, Y.~Adi, A.~Lee, and P.-J. Chen, ``A holistic cascade system, benchmark, and human evaluation protocol for expressive speech-to-speech translation,'' 2023.

\bibitem{gujarathi2021review}
P.~Gujarathi and S.~R. Patil, ``Review on unit selection-based concatenation approach in text to speech synthesis system,'' in \emph{Cybernetics, Cognition and Machine Learning Applications}, ser. Algorithms for Intelligent Systems (AIS), 2021, pp. 191--202.

\bibitem{parametric_tts_2011}
S.~King, ``An introduction to statistical parametric speech synthesis,'' \emph{c Indian Academy of Sciences}, vol.~36, pp. 837--852, 11 2011.

\bibitem{deeplearning_ex1}
Y.~Lecun and Y.~Bengio, ``Convolutional networks for images, speech, and time-series,'' in \emph{Convolutional Networks for Images, Speech, and Time-Series}, 01 1995.

\bibitem{deeplearning_ex2}
Z.~Fayyaz, D.~Platnick, H.~Fayyaz, and N.~Farsad, ``Deep unfolding for iterative stripe noise removal,'' in \emph{2022 International Joint Conference on Neural Networks (IJCNN)}, 2022, pp. 1--7.

\bibitem{deeplearning_ex3}
D.~Platnick, S.~Khanzadeh, A.~Sadeghian, and R.~Valenzano, ``Gansemble for {Small} and {Imbalanced} {Data} {Sets}: A {Baseline} for {Synthetic} {Microplastics} {Data},'' \emph{Proceedings of the Canadian Conference on Artificial Intelligence}, may 27 2024, https://caiac.pubpub.org/pub/0hhra7j6.

\bibitem{deeplearning_ex4}
\BIBentryALTinterwordspacing
Y.~Ning, S.~He, Z.~Wu, C.~Xing, and L.-J. Zhang, ``A review of deep learning based speech synthesis,'' \emph{Applied Sciences}, vol.~9, no.~19, 2019. [Online]. Available: \url{https://www.mdpi.com/2076-3417/9/19/4050}
\BIBentrySTDinterwordspacing

\bibitem{deep-learning2024}
H.~Barakat, O.~Turk, and C.~Demiroglu, ``Deep learning-based expressive speech synthesis: a systematic review of approaches, challenges, and resources,'' \emph{EURASIP Journal on Audio, Speech, and Music Processing}, vol.~11, p.~11, 2024.

\bibitem{oord2016wavenet}
A.~van~den Oord, S.~Dieleman, H.~Zen, K.~Simonyan, O.~Vinyals, A.~Graves, N.~Kalchbrenner, A.~Senior, and K.~Kavukcuoglu, ``Wavenet: A generative model for raw audio,'' 2016.

\bibitem{tacotron1_2022}
\BIBentryALTinterwordspacing
Y.~Wang, R.~J. Skerry{-}Ryan, D.~Stanton, Y.~Wu, R.~J. Weiss, N.~Jaitly, Z.~Yang, Y.~Xiao, Z.~Chen, S.~Bengio, Q.~V. Le, Y.~Agiomyrgiannakis, R.~Clark, and R.~A. Saurous, ``Tacotron: {A} fully end-to-end text-to-speech synthesis model,'' \emph{CoRR}, vol. abs/1703.10135, 2017. [Online]. Available: \url{http://arxiv.org/abs/1703.10135}
\BIBentrySTDinterwordspacing

\bibitem{vallex_2023}
Z.~Zhang, L.~Zhou, C.~Wang, S.~Chen, Y.~Wu, S.~Liu, Z.~Chen, Y.~Liu, H.~Wang, J.~Li, L.~He, S.~Zhao, and F.~Wei, ``Speak foreign languages with your own voice: Cross-lingual neural codec language modeling,'' 2023.

\bibitem{xtts}
\BIBentryALTinterwordspacing
G.~Eren and T.~C.~T. Team, ``Coqui tts,'' dec 2023. [Online]. Available: \url{https://doi.org/10.5281/zenodo.10363832}
\BIBentrySTDinterwordspacing

\bibitem{data_dependent_models_2017}
\BIBentryALTinterwordspacing
I.~Rebai, Y.~BenAyed, W.~Mahdi, and J.-P. Lorré, ``Improving speech recognition using data augmentation and acoustic model fusion,'' \emph{Procedia Computer Science}, vol. 112, pp. 316--322, 2017, knowledge-Based and Intelligent Information \& Engineering Systems: Proceedings of the 21st International Conference, KES-20176-8 September 2017, Marseille, France. [Online]. Available: \url{https://www.sciencedirect.com/science/article/pii/S187705091731342X}
\BIBentrySTDinterwordspacing

\bibitem{bus_dispatch_2001}
J.~Strathman, T.~Kimpel, K.~Dueker, R.~Gerhart, K.~Turner, D.~Griffin, S.~Callas, W.~Ahmad, T.~Curtis, W.~Martin, J.~Alldrin, R.~Mcalister, B.~Shields, K.~Saviers, B.~Bradford, M.~Seid, T.~Ferguson, and L.~Wilkinson, ``Bus transit operations control: Review and an experiment involving tri-met's automated bus dispatching system,'' \emph{Journal of Public Trans.}, vol.~4, 12 2001.

\bibitem{medical_alert_2013}
O.~Eyesan and S.~Okuboyejo, ``Design and implementation of a voice-based medical alert system for medication adherence,'' in \emph{Design and Implementation of a Voice-Based Medical Alert System for Medication Adherence}, vol.~9, 10 2013.

\bibitem{airport_alert_2019}
P.~Samaras and M.~Ferreira, ``Emergency communication systems effectiveness in an airport environment,'' \emph{Journal of business continuity \& emergency planning}, vol.~12, pp. 242--252, 01 2019.

\bibitem{SER_gender_dependent_nn_2023}
\BIBentryALTinterwordspacing
V.~Singh and S.~Prasad, ``Speech emotion recognition system using gender dependent convolution neural network,'' \emph{Procedia Computer Science}, vol. 218, pp. 2533--2540, 2023, international Conference on Machine Learning and Data Engineering. [Online]. Available: \url{https://www.sciencedirect.com/science/article/pii/S1877050923002272}
\BIBentrySTDinterwordspacing

\bibitem{RAVDESS_dataset}
\BIBentryALTinterwordspacing
S.~R. Livingstone and F.~A. Russo, ``{The Ryerson Audio-Visual Database of Emotional Speech and Song (RAVDESS)},'' Apr. 2018. [Online]. Available: \url{https://doi.org/10.5281/zenodo.1188976}
\BIBentrySTDinterwordspacing

\bibitem{CREMAD_dataset}
H.~Cao, D.~G. Cooper, M.~K. Keutmann, R.~C. Gur, A.~Nenkova, and R.~Verma, ``Crema-d: Crowd-sourced emotional multimodal actors dataset,'' \emph{IEEE Transactions on Affective Computing}, vol.~5, no.~4, pp. 377--390, 2014.

\bibitem{SAVEE_dataset}
O.~C. Phukan, A.~B. Buduru, and R.~Sharma, ``A comparative study of pre-trained speech and audio embeddings for speech emotion recognition,'' 2023.

\bibitem{SP2/E8H2MF_2020}
\BIBentryALTinterwordspacing
M.~K. Pichora-Fuller and K.~Dupuis, ``{Toronto emotional speech set (TESS)},'' 2020. [Online]. Available: \url{https://doi.org/10.5683/SP2/E8H2MF}
\BIBentrySTDinterwordspacing

\bibitem{pitch_loudness}
\BIBentryALTinterwordspacing
J.~Za{\"{\i}}di, H.~Seut{\'{e}}, B.~van Niekerk, and M.~Carbonneau, ``Daft-exprt: Robust prosody transfer across speakers for expressive speech synthesis,'' \emph{CoRR}, vol. abs/2108.02271, 2021. [Online]. Available: \url{https://arxiv.org/abs/2108.02271}
\BIBentrySTDinterwordspacing

\bibitem{imagenet}
J.~Deng, W.~Dong, R.~Socher, L.-J. Li, K.~Li, and L.~Fei-Fei, ``Imagenet: A large-scale hierarchical image database,'' in \emph{2009 IEEE Conference on Computer Vision and Pattern Recognition}, 2009, pp. 248--255.

\bibitem{resnet50_images}
\BIBentryALTinterwordspacing
H.~Sinha, V.~Awasthi, and P.~K. Ajmera, ``Audio classification using braided convolutional neural networks,'' \emph{IET Signal Processing}, vol.~14, no.~7, pp. 448--454, 2020. [Online]. Available: \url{https://ietresearch.onlinelibrary.wiley.com/doi/abs/10.1049/iet-spr.2019.0381}
\BIBentrySTDinterwordspacing

\bibitem{adam_op_2017}
D.~P. Kingma and J.~Ba, ``Adam: A method for stochastic optimization,'' 2017.

\bibitem{communication2023seamlessm4t}
S.~Communication, L.~Barrault, Y.-A. Chung, M.~C. Meglioli, D.~Dale, N.~Dong, P.-A. Duquenne, H.~Elsahar, H.~Gong, K.~Heffernan, J.~Hoffman, C.~Klaiber, P.~Li, D.~Licht, J.~Maillard, A.~Rakotoarison, K.~R. Sadagopan, G.~Wenzek, E.~Ye, B.~Akula, P.-J. Chen, N.~E. Hachem, B.~Ellis, G.~M. Gonzalez, J.~Haaheim, P.~Hansanti, R.~Howes, B.~Huang, M.-J. Hwang, H.~Inaguma, S.~Jain, E.~Kalbassi, A.~Kallet, I.~Kulikov, J.~Lam, D.~Li, X.~Ma, R.~Mavlyutov, B.~Peloquin, M.~Ramadan, A.~Ramakrishnan, A.~Sun, K.~Tran, T.~Tran, I.~Tufanov, V.~Vogeti, C.~Wood, Y.~Yang, B.~Yu, P.~Andrews, C.~Balioglu, M.~R. Costa-jussà, O.~Celebi, M.~Elbayad, C.~Gao, F.~Guzmán, J.~Kao, A.~Lee, A.~Mourachko, J.~Pino, S.~Popuri, C.~Ropers, S.~Saleem, H.~Schwenk, P.~Tomasello, C.~Wang, J.~Wang, and S.~Wang, ``Seamlessm4t: Massively multilingual \& multimodal machine translation,'' 2023.

\bibitem{NISQA_2021}
\BIBentryALTinterwordspacing
G.~Mittag, B.~Naderi, A.~Chehadi, and S.~Möller, ``Nisqa: A deep cnn-self-attention model for multidimensional speech quality prediction with crowdsourced datasets,'' in \emph{Interspeech 2021}.\hskip 1em plus 0.5em minus 0.4em\relax ISCA, Aug. 2021. [Online]. Available: \url{http://dx.doi.org/10.21437/Interspeech.2021-299}
\BIBentrySTDinterwordspacing

\bibitem{NISQA2_2022}
G.~Yi, W.~Xiao, Y.~Xiao, B.~Naderi, S.~Möller, W.~Wardah, G.~Mittag, R.~Cutler, Z.~Zhang, D.~S. Williamson, F.~Chen, F.~Yang, and S.~Shang, ``Conferencingspeech 2022 challenge: Non-intrusive objective speech quality assessment (nisqa) challenge for online conferencing applications,'' 2022.

\bibitem{cafe}
\BIBentryALTinterwordspacing
P.~Gournay, O.~Lahaie, and R.~Lefebvre, ``A canadian french emotional speech dataset,'' Nov. 2018. [Online]. Available: \url{https://doi.org/10.5281/zenodo.1478765}
\BIBentrySTDinterwordspacing

\bibitem{emodb}
F.~Burkhardt, A.~Paeschke, M.~Rolfes, W.~F. Sendlmeier, and B.~Weiss, ``A database of {German} emotional speech.'' in \emph{Proceedings of the Annual Conference of the International Speech Communication Association (INTERSPEECH)}, vol.~5.\hskip 1em plus 0.5em minus 0.4em\relax Lisbon, Portugal: ISCA, 2005, pp. 1517--1520.

\bibitem{openvoice_2024}
Z.~Qin, W.~Zhao, X.~Yu, and X.~Sun, ``Openvoice: Versatile instant voice cloning,'' 2024.

\bibitem{vits}
J.~Kim, J.~Kong, and J.~Son, ``Conditional variational autoencoder with adversarial learning for end-to-end text-to-speech,'' 2021.

\bibitem{styletts}
Y.~A. Li, C.~Han, V.~S. Raghavan, G.~Mischler, and N.~Mesgarani, ``Styletts 2: Towards human-level text-to-speech through style diffusion and adversarial training with large speech language models,'' 2023.

\bibitem{devlin2019bert}
J.~Devlin, M.-W. Chang, K.~Lee, and K.~Toutanova, ``Bert: Pre-training of deep bidirectional transformers for language understanding,'' 2019.

\bibitem{babu2021xlsr}
A.~Babu, C.~Wang, A.~Tjandra, K.~Lakhotia, Q.~Xu, N.~Goyal, K.~Singh, P.~von Platen, Y.~Saraf, J.~Pino, A.~Baevski, A.~Conneau, and M.~Auli, ``Xls-r: Self-supervised cross-lingual speech representation learning at scale,'' 2021.

\end{thebibliography}

\vspace{12pt}

\end{document}